\def\year{2017}\relax
\documentclass[letterpaper]{article}
\pdfoutput=1
\usepackage{aaai17}
\usepackage{times}
\usepackage{helvet}
\usepackage{courier}

\usepackage{graphicx}
\usepackage{amsmath}
\usepackage{amssymb}
\usepackage{makecell}
\usepackage{color}

\begin{document}
%
\title{Text-guided Attention Model for Image Captioning}
\author{Jonghwan Mun, Minsu Cho, Bohyung Han\\
Department of Computer Science and Engineering, POSTECH, Korea \\
\{choco1916, mscho, bhhan\}@postech.ac.kr}

\newcommand{\etal}{\textit{et al}.}
\newcommand{\ie}{\textit{i}.\textit{e}., }
\newcommand{\eg}{\textit{e}.\textit{g}.}

\nocopyright
\maketitle

\begin{abstract}
Visual attention plays an important role to understand images and demonstrates its effectiveness in generating natural language descriptions of images.
On the other hand, recent studies show that language associated with an image can steer visual attention in the scene during our cognitive process.  
Inspired by this, we introduce a text-guided attention model for image captioning, which learns to drive visual attention using associated captions.
For this model, we propose an exemplar-based learning approach that retrieves from training data associated captions with each image, and use them to learn attention on visual features. 
Our attention model enables to describe a detailed state of scenes by distinguishing small or confusable objects effectively.
We validate our model on MS-COCO Captioning benchmark and achieve the state-of-the-art performance in standard metrics.
\end{abstract}

\section{Introduction}
\label{sec:introduction}

Image captioning, which aims at automatically generating descriptions of an image in natural language, is one of the major problems in scene understanding. 
This problem has long been viewed as an extremely challenging task because it requires to capture and express low- to high-level aspects of local and global areas in an image containing a variety of scene elements as well as their relationships.

Despite such challenges, the problem has drawn wide attention and significant improvement has been achieved over the past few years through advance of deep neural networks together with construction of large-scale datasets.
For example, the encoder-decoder framework~\cite{kiros2014unifying,mao2014deep,vinyals2015show,CVPR16what} formulates the image captioning problem using a deep neural network trained end-to-end rather than a pipeline of separate subtasks such as detection of objects and activities, ordering words, etc.

\begin{figure}[t]
	\centering
	\scalebox{0.95}{
		\begin{tabular}{c}
		\makecell[{c}]{ \hspace{-0.35cm}\includegraphics[width=\linewidth]{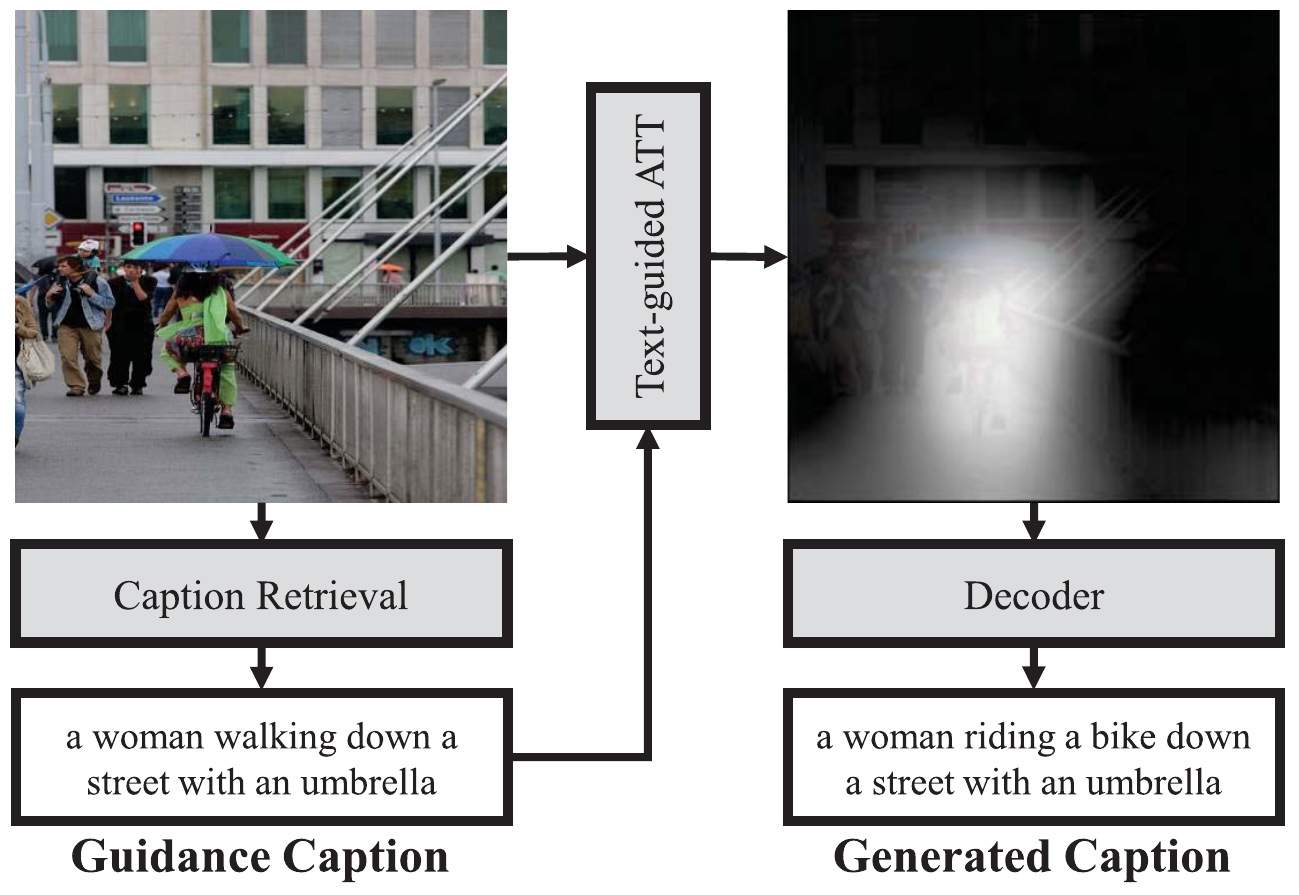} } 
		\end{tabular}
	}
	\caption{
	An overview of the proposed framework.
	In our approach, the caption associated with an image guides visual attention to better focus on relevant regions in the scene, resulting in proper and fine-grained image captions even from cluttered scenes.
	}
	\label{fig:T-ATT}
\end{figure} 

Recently, one of major advances in image captioning has been made by incorporating a form of visual attention into the image encoder~\cite{Xu2015show,you2016image}. 
Instead of compressing an entire image into a flat representation, attention models allow encoder to focus only on relevant features for image captioning. This helps to describe fine details in regions of interest even from a cluttered image.

On the other hand, recent studies in psychology reveal close relationships between language and visual attention~\cite{mishrainteraction}.
Spoken or text language that expresses objects and events leads a person to construct a mental representation about them.
This representation contributes to anticipatory eye movements of the person, thus resulting in successful detection of objects and actions~\cite{lupyan2010correction}.
This suggests that spoken or text language associated with a scene can provide useful information for visual attention in the scene.

In this paper, we introduce a text-guided attention model for image captioning, which learns visual attention from associated exemplar captions for a given image, referred to as {\em guidance captions}, and enables to generate proper and fine-grained captions for the image.
In our approach, we extract the guidance captions by retrieving visually similar images in training dataset, which would share regions of interest with the given image.
The guidance captions steer our attention model to focus only on relevant regions for image captioning.
Figure \ref{fig:T-ATT} illustrates the process with an example result.
The associated caption successfully guides the model to attend to important regions, and helps it generate detailed and suitable captions.

The main contributions of this paper are as follows:
\begin{itemize}
	\item 
		We introduce a novel attention model for image captioning, referred to as {\em text-guided attention model}, which directly exploits exemplar captions in training data as a guidance source for visual attention. This exemplar-based attention model is learned within our image captioning architecture in an end-to-end manner.
	\item 
		We develop a sampling-based scheme to learn attention using multiple exemplar guidance captions. 
		This avoids overfitting in training and alleviates the problem of learning misleading attention from noisy guidance captions.
	\item
		Our image captioning network with the proposed attention model achieves the state-of-the-art performance on MS-COCO Captioning benchmark~\cite{lin2014microsoft}, outperforming recent attention-based methods. 
\end{itemize}

The rest of this paper is organized as follows.
We first review the related work and overview our architecture. 
Then, each component of our method is described in details, and training and inference procedures are discussed.
Finally, we present experimental results and compare our algorithm with the existing ones.

\section{Related Work}
\label{sec:related_work}

We briefly review existing algorithms in the context of image captioning and visual attention. 

\subsection{Image Captioning} 
Most of recent methods for image captioning are based on the encoder-decoder framework~\cite{kiros2014unifying,mao2014deep,vinyals2015show,CVPR16what}.  
This approach encodes an image using CNN and transforms the encoded representation into a caption using RNN.
\cite{vinyals2015show} introduces a basic encoder-decoder model that extracts an image feature using GoogLeNet~\cite{szegedy2015going} and feeds the feature as the first word into the Long Short Term Memory (LSTM) decoder. 
Instead of feeding the image feature into RNN, \cite{mao2014deep} proposes a technique to transform the image feature and the hidden state of RNN into an intermediate embedding space. Each word is predicted based on the aggregation of the embedded features at each time step.  
On the other hand, image attributes are learned and the presence of the attributes, instead of an encoded image feature, is given to LSTM as an input to generate captions~\cite{CVPR16what}.
In \cite{kiros2014unifying}, multi-modal space between an image and a caption is learned, and an embedded image feature is fed to a language model as visual information.

\subsection{Attention Model} 
Inspired by the perceptual process of human being, attention mechanisms have been developed for a variety of tasks such as object recognition~\cite{ba2014multiple,kantorov2016}, image generation~\cite{gregor2015draw}, semantic segmentation~\cite{hong2016learning} and visual question answering~\cite{andreas2015deep,noh2016training,xu2015ask,yang2015stacked}.
In image captioning, there are two representative methods adopting a form of visual attention.
Spatial attention for a word is estimated using the hidden state of LSTM at each time step in \cite{Xu2015show} while the algorithm in \cite{you2016image} computes a semantic attention on word candidates for caption, uses it to refine input and output of LSTM at each step.
While these methods demonstrate the effectiveness of visual attention in image captioning,  they do not consider a more direct use of captions available in training data.
Our method leverages the captions in both learning and inference, exploiting them as high-level guidance for visual attention.
To the best of our knowledge, the proposed method is the first work for image captioning that combines visual attention with a guidance of associated text language.

\begin{figure*}[t]
	\centering
	\scalebox{0.95}{
		\includegraphics[width=1\textwidth]{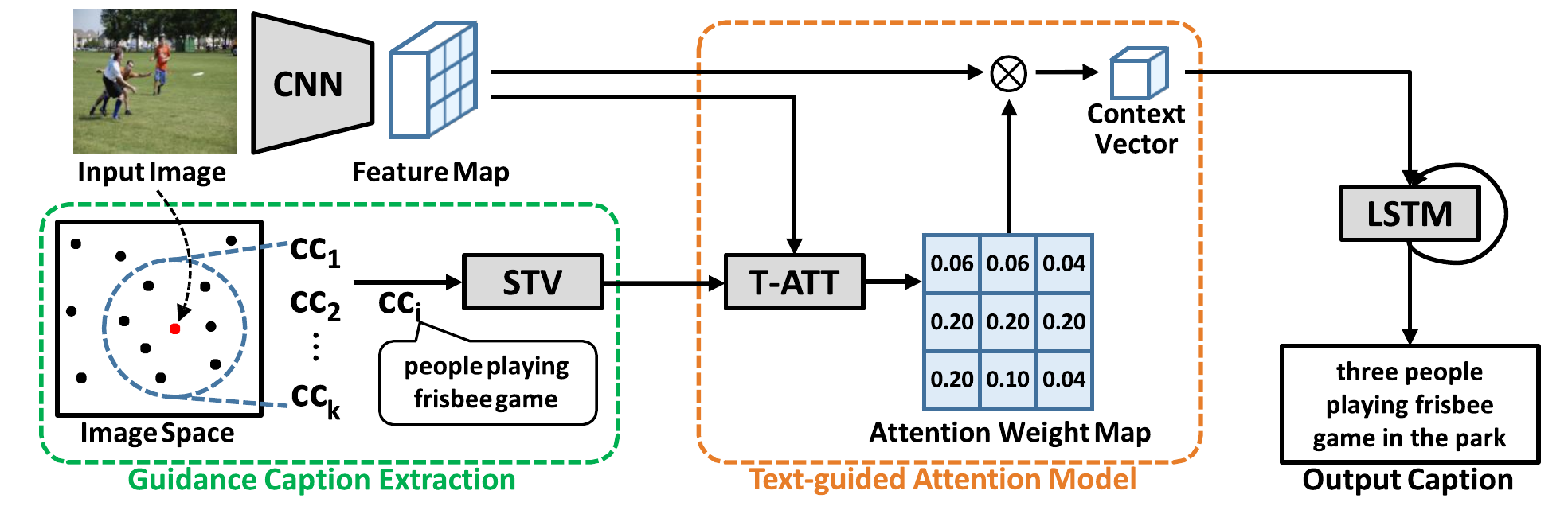}
	}
	\caption{
	Overall architecture for image captioning with text-guided attention. 
	Given an input image,  we first retrieve top $k$ candidate captions ($\rm{CC}$) from training data using both visual similarity and caption consensus scores. We randomly select one among them as the guidance caption in training while using all candidates as guidance captions in testing time.
	The text-guided attention layer (T-ATT) computes an attention weight map where regions relevant to the given guidance caption have higher attention weights.
	A context vector is obtained by aggregating image feature vectors weighted by the attention map.
	Finally, the LSTM decoder generates an output caption from the context vector. For the details, see text. 
	}
	\label{fig:architecture}
\end{figure*}

\section{Overview}
\label{sec:overview}

For a pair of an image $I$ and a caption $c$ consisting of $T$ words ($w_1,w_2,...,w_T$), conventional image captioning approaches with an attention model~\cite{Xu2015show} minimize a negative log-likelihood in training as follows:
\begin{align}
	\mathcal{L} &={-\log p(c|I)} \nonumber \\
			     &=\sum_{t=0}^{T}{-\log p(w_{t+1}|w_{t},h_{t-1},f_{att}(I,h_{t-1}))},
\label{eq:log-likelihood}
\end{align}
where $f_{att}$ is the attention model and $h_{t-1}$ is the previous hidden state of LSTM.
Note that we use two additional words in Eq~\eqref{eq:log-likelihood}: $w_0$ ($<$BOS$>$) and $w_{T+1}$ ($<$EOS$>$) indicating the begin and end of a sentence, respectively.
The attention is adaptively computed at each time step based on the previous hidden state.

In our approach, we leverage a guidance caption $s$ as associated text language to steer visual attention. Our image captioning network is trained to minimize the following loss:
\begin{align}
	\mathcal{L}&={-\log p(c|f_{T\textnormal{-}att}(I,s))} \nonumber \\
			    &=-\log p(w_{1}|w_{0}, f_{T\textnormal{-}att}(I,s)) \nonumber \\
			    &\quad+\sum_{t=1}^{T}{-\log p(w_{t+1}| w_{t}, h_{t-1})},
\label{eq:loss}
\end{align}
where ${f_{T\textnormal{-}att}}$ is the proposed text-guided attention model.
Contrary to previous attention-based approaches, the attention of our model is driven not only by image features but also by text features obtained from the guidance captions.
The attention is computed only once at the beginning and used to compute the initial hidden state $h_{-1}$.

The overall architecture is illustrated in Figure \ref{fig:architecture}.
For a given image, we first retrieve visually similar images in training dataset and select a guidance caption. Then, the image and the guidance caption are transformed to separate multi-dimensional feature vectors, which are used to compute an attention weight map for the image.
In the attention map, the regions relevant to the guidance caption are highlighted while irrelevant regions are suppressed.
The decoder generates an output caption based on a weighted sum of the image feature vectors, where the weights are determined by the attention map.

\section{Algorithm and Architecture}
\label{sec:architecture}
This section discusses the details of our algorithm and the proposed architecture.

\subsection{Guidance Caption Extraction}

Ground-truth captions would be ideal as guidance caption for visual attention, but unavailable during inference for image captioning. 
Therefore, we use an exemplar-based approach to obtaining guidance captions that provide useful information for visual attention.
We observe that visually similar images tend to share the salient objects and events that are often described in their captions.
Based on this observation, we use the consensus captions~\cite{devlin2015exploring} for guidance caption extraction.
The consensus caption is defined as a representative caption consented by the captions of $n$ nearest neighbors in a visual feature space.
The caption is likely to coincide with the caption of the query image than that of the nearest neighbor in practice.
For each image, we collect a set of captions $C_{\rm NN}$ from $n$ nearest neighbor images in a visual feature space.
The consensus score $s_i$ for a caption $c_i$ is computed by average similarity to all the other captions in $C_{\rm NN}$:
\begin{equation}
	s_i=\frac{1}{|C_{\rm NN}|-1}\sum_{c' \in {C_{\rm NN}} \backslash \{ c_i \}} \mathrm{Sim}(c_i,c'),\label{eq:cons_cap}
\end{equation}
where $\mathrm{Sim}(c_i,c')$ is the similarity between two captions $c_i$ and $c'$.
We use CIDEr~\cite{vedantam2015cider} as the similarity function.
We maintain a set of top $k$ captions $(k<|C_{\rm NN}|)$ in terms of the consensus score as guidance caption candidates rather than only a top one. 

\subsection{Encoder}

There exist two encoders in our architecture, an image encoder and a guidance caption encoder.
We employ a CNN and an RNN for the image and the guidance caption encoder, respectively.
The CNN image encoder extracts feature map $\mathbf{f}_I\in\mathbb{R}^{\emph{R}\times\emph{R} \times \emph{P}}$, which contains $P$-dimensional features of corresponding receptive fields.
For the RNN guidance caption encoder, we employ the Skip-Thought Vector model (STV)~\cite{kiros2015skip}.
This sentence embedding model is trained in an unsupervised learning technique by predicting surrounding two sentences on a large corpus more than 74M sentences.
The STV is composed of Gated Recurrent Units (GRU)~\cite{chung2014empirical}, which is similar to but simpler than LSTM, and the guidance caption feature $\mathbf{f}_S\in\mathbb{R}^{\emph{S}}$ is obtained from the last hidden state of GRU.
Note that we fix the parameters of the two encoders during training.

\subsection{Text-guided Attention Model}

Given image feature $\mathbf{f}_I$ and guidance caption feature $\mathbf{f}_S$, our next goal is to extract a context vector that describes relevant content of the image.
Let us denote a region feature of $i$th location in image by $\mathbf{f}_{I}^{i}$.
Our attention model computes an attention weight of $\mathbf{f}_I^i$ for the guidance caption feature $\mathbf{f}_S$ using a simple feedforward neural network:
\begin{align}
	\mathbf{e}_i&=\mathbf{W}_{att}(\mathbf{W}_{I}\mathbf{f}_I^i+\mathbf{W}_{S}\mathbf{f}_S), \\
	\mathbf{\alpha}_{i}&=\mathrm{Softmax}({\mathbf{e}_i),\label{eq:att_weight}}
\end{align}
where $\mathbf{W}_{I}, \mathbf{W}_{S}, \mathbf{W}_{att}$ are learned embedding matrices for image feature, guidance caption feature, and their weighted sum, respectively. 
After calculating attention weights for all regions, softmax is applied so that the sum of attention weight becomes one.
Then, we obtain the context vector $\mathbf{z}\in\mathbb{R}^{\emph{P}}$ as a weighted sum of the image feature vectors:
\begin{align}
	\mathbf{z}=\sum_{i}\mathbf{\alpha}_i\mathbf{f}_I^i.
\end{align}

In training the attention model, we regularize the attention weight in (\ref{eq:att_weight}) by adding an entropy term $\mathbf{\alpha}\log (\mathbf{\alpha})$ into (\ref{eq:loss}), which encourages attention weights to be uniform and penalizes excessive attention to a certain region.
During training, the attention model gradually learns the regions of interest starting from uniform attention.

\subsection{Decoder}

The overall decoder consists of a word embedding layer, a LSTM unit, and a word prediction layer.
Let us assume that the caption is composed of $T$ words ($w_1,w_2,...,w_{T}$).
For a beginning of the sentence, we add the word $w_0$, that is $<$BOS$>$.
Our decoder is formulated by
\begin{align}
	\mathbf{x}_{-1} &= \mathbf{W}_{z}\mathbf{z}, \\
	\mathbf{x}_t &= \mathbf{W}_{e}w_{t}, \\ 
	\mathbf{h}_t &= \mathrm{LSTM}(\mathbf{x}_t,\mathbf{h}_{t-1}), \\
	\mathbf{p}_{t+1} &= \mathrm{Softmax}({\mathbf{W}_{h}}\mathbf{h}_{t}),
\end{align}
where $\mathbf{W}_z$, $\mathbf{W}_e$ and $\mathbf{W}_h$ are learned embedding matrices for the context vector, the input word and the hidden state.
$\mathbf{p}_{t+1}$ indicates a probability distribution over all words.
At each time step $t$, the input word $\mathbf{w}_t$ is projected into the word vector space.
The LSTM unit computes the current hidden state $\mathbf{h}_t$ based on the word vector $\mathbf{x}_t$ and the previous hidden state $\mathbf{h}_{t-1}$. 
Then, the word is predicted based on the current hidden state $\mathbf{h}_t$.
The predicted word and the current hidden state is fed back into the decoder unit in the next time step and the entire process is repeated until emitting the word $<$EOS$>$.
At the initial step ($t$=$-1$), the context vector containing the image information is used as an initial word.
Here, the context vector is embedded to match its dimension with that of the word vector.
Given the embedded context vector $\mathbf{x}_{-1}$, LSTM unit computes the initial hidden state $\mathbf{h}_{-1}$ using a vector with all zeros in the place of the previous hidden state $h_{-2}$.
The word embedding and prediction layers share their weights to reduce the number of parameters~\cite{mao2015learning}.

\section{Learning and Inference}
\label{sec:train_inference}

\begin{table*}[t]
	\centering
	\caption{
		Performance on MS-COCO test split. 
		$\diamond$ means model ensemble (single-model results are not reported) and $\dagger$ stands for using domain-specific pretrained encoder.
		Numbers in red and blue denote the best and second-best algorithms, respectively.
	}
	\scalebox{0.85}{
		\begin{tabular}{l|cccccc}
			& ~~BLEU-1~~ & ~~BLEU-2~~ & ~~BLEU-3~~ & ~~BLEU-4~~ & ~~METEOR~~ & ~~CIDEr~~ \\ \hline
			NIC~\cite{vinyals2015show}$^{\diamond}$		&\textendash	&\textendash	&\textendash	&0.277	&0.237	&0.855	\\
			m-RNN~\cite{mao2014deep}$^{}$			&0.686	&0.511	&0.375	&0.280	&0.228	&0.842	\\
			LRCN~\cite{donahue2015long}$^{}$			&0.669	&0.489	&0.349	&0.249	&\textendash	&\textendash	\\
			MSR~\cite{fang2015captions}$^{\dagger}$		&\textendash	&\textendash	&\textendash	&0.257	&0.236	&\textendash	\\
			Soft-ATT~\cite{Xu2015show}$^{}$			&0.707	&0.492	&0.344	&0.243	&0.239	&\textendash	\\ 
			Hard-ATT~\cite{Xu2015show}$^{}$			&0.718	&0.504	&0.357	&0.250	&0.230	&\textendash	\\ 
			Semantic ATT~\cite{you2016image}$^{\dagger\diamond}$	&0.709	&0.537	&0.402	&0.304	&0.243	&\textendash	\\ 
			Attribute-LSTM~\cite{CVPR16what}$^{\dagger}$	&{\color{blue}\textbf{0.740}}&0.560	&0.420	&0.310	&{\color{red}\textbf{0.260}}	&0.940	\\ \hline
			mCC										&0.670	&0.486	&0.345	&0.244	&0.225	&0.791	\\
			Ours-Uniform	(VGG-FCN)					&0.733	&0.563	&0.420	&0.313	&0.249	&0.968	\\
			Ours-ATT-mCC (VGG-FCN)					&0.732	&0.563	&0.421	&0.313	&0.250	&0.972	\\ 
			Ours-ATT-kCC (VGG-FCN)					&0.735	&{\color{blue}\textbf{0.566}}&{\color{blue}\textbf{0.424}}&{\color{blue}\textbf{0.316}}&0.251	&{\color{blue}\textbf{0.982}}	\\  \hline
			Ours-Uniform (ResNet)						&0.741	&0.572	&0.429	&0.319	&0.253	&0.996	\\ 
			Ours-ATT-mCC (ResNet)						&0.743	&0.575	&0.432	&0.323	&0.255	&1.010	\\
			Ours-ATT-kCC (ResNet)						& {\color{red} \textbf{0.749}} & {\color{red} \textbf{0.581}} & {\color{red} \textbf{0.437}} & {\color{red} \textbf{0.326}} 	& 0.257 & {\color{red} \textbf{1.024}}	\\ \hline
		\end{tabular}
	}
	\label{table:coco-val}
\end{table*}

\begin{table*}[th]
	\centering
	\caption{
		Performance comparison with the state-of-the-art results on MS-COCO Image Captioning Challenge.
		$\diamond$ means model ensemble (single-model results are not reported) and $\dagger$ stands for using domain-specific pretrained or fine-tuned encoder.
		Numbers in red and blue denote the best and second-best algorithms, respectively.
	}
	\scalebox{0.85}{
		\begin{tabular}{l|cc|cc|cc|cc|cc|cc}
			& \multicolumn{2}{c|}{BLEU-1}	& \multicolumn{2}{c|}{BLEU-2}	& \multicolumn{2}{c|}{BLEU-3}	& \multicolumn{2}{c|}{BLEU-4}	&\multicolumn{2}{c|}{METEOR}	& \multicolumn{2}{c}{CIDEr}	\\ 
			 & c5 & c40 & c5 & c40 & c5 & c40 & c5 & c40 & c5 & c40 & c5 & c40 \\ \hline
			NIC~\cite{vinyals2015show}$^{\dagger\diamond}$	& 0.713 & 0.895 & 0.542 & 0.802 & 0.407 & 0.694 & 0.309 & 0.587 & {\color{blue}\textbf{0.254}} &{\color{red}\textbf{0.346}} & {\color{blue}\textbf{0.943}} & 0.946	\\
			m-RNN~\cite{mao2014deep}						& 0.716 & 0.890 & 0.545 & 0.798 & 0.404 & 0.687 & 0.299 & 0.575 & 0.242 & 0.325 & 0.917 & 0.935	\\
			LRCN~\cite{donahue2015long}$^{\dagger}$		& 0.718 & 0.895 & 0.548 & 0.804 & 0.409 & 0.695 & 0.306 & 0.585 & 0.247 & 0.335 & 0.921 & 0.934	\\
			MSR~\cite{fang2015captions}$^{\dagger}$			& 0.715 & {\color{blue}\textbf{0.907}} & 0.543 & {\color{blue}\textbf{0.819}} & 0.407 & {\color{blue}\textbf{0.710}} & 0.308 & 0.601 & 0.248 & 0.339 & 0.931 & 0.937	\\ 
			Hard-ATT~\cite{Xu2015show}							& 0.705 & 0.881 & 0.528 & 0.779 & 0.383 & 0.658 & 0.277 & 0.537 & 0.241 & 0.322 & 0.865 & 0.893	\\
			Semantic ATT~\cite{you2016image}$^{\dagger\diamond}$	& {\color{blue}\textbf{0.731}} & 0.900  & {\color{blue}\textbf{0.565}} & 0.815 & {\color{blue}\textbf{0.424}} & 0.709 & {\color{blue}\textbf{0.316}} & {\color{blue}\textbf{0.599}} & 0.250 & 0.335 & {\color{blue}\textbf{0.943}} & {\color{blue}\textbf{0.958}}	\\
			Attribute-LSTM~\cite{CVPR16what}$^{\dagger}$	& 0.725 & 0.892 & 0.556 & 0.803 & 0.414 & 0.694	& 0.306 & 0.582 & 0.246 & 0.329 & 0.911 & 0.924 	\\ \hline
			Ours-ATT-mCC (VGG-FCN)			& 0.727 & 0.901 & 0.558 & 0.812 & 0.414 & 0.701	& 0.305 & 0.585 & 0.247 & 0.330 & 0.930 & 0.942 	\\ 
			Ours-ATT-kCC (VGG-FCN)			& {\color{blue}\textbf{0.731}} & 0.902 & 0.560 & 0.814 & 0.416 & 0.703 & 0.307 & 0.587 & 0.248 & 0.331 & 0.938 & 0.951 	\\ 
			Ours-ATT-mCC (ResNet)				& 0.739 & 0.913 & 0.570 & 0.828 & 0.427 & 0.719	& 0.318 & 0.603 & 0.253 & 0.339 & 0.972 & 0.988 	\\ 
			Ours-ATT-kCC (ResNet)				& {\color{red} \textbf{0.743}} & {\color{red} \textbf{0.915}} & {\color{red} \textbf{0.575}} & {\color{red} \textbf{0.832}} & {\color{red} \textbf{0.431}} &{\color{red} \textbf{0.722}} & {\color{red} \textbf{0.321}} & {\color{red} \textbf{0.607}} & {\color{red} \textbf{0.255}} &0.341 & {\color{red} \textbf{0.987}} & {\color{red} \textbf{1.001}}   \\ \hline
		\end{tabular}
	}
	\label{table:coco-test}
\end{table*}

In our approach, the quality of guidance captions is crucial for both training and testing. 
Even though we manage to filter out many noisy guidance captions by the consensus caption scheme of Eq.~\eqref{eq:cons_cap}, we observe that our attention model often suffers from overfitting when using a caption of the highest consensus score as the guidance caption.
To tackle this issue, we use a set of top $k$ consensus captions as candidates of the guidance caption in both training and testing.

In training, we randomly select a caption among the top $k$ consensus captions for each image. This random sampling strategy allows our attention model to avoid overfitting and learn proper attention from diverse consensus captions.

In testing, we generate $k$ captions using all the $k$ consensus captions as guidance captions, and then select the best one by reranking the $k$ captions. 
Note that we apply the beam search of size 2 while generating a caption with each consensus caption as the guidance caption.
For the reranking, we adopt the scheme of~\cite{mao2014deep} that measures CIDEr similarity with captions of its $n$ nearest neighbor images in the training data. 
In our experiments, we fix $n=60$ and $k=10$ in both training and testing.

\begin{figure*}[t]
	\centering
	\scalebox{0.70} {
		\begin{tabular}{ccc}
		 \thead{ \includegraphics[width=0.22\textwidth]{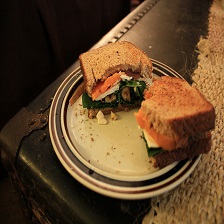}  \includegraphics[width=0.22\textwidth]{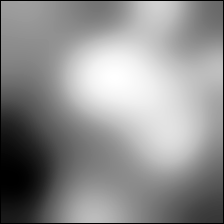} }	& 
		\thead{ \includegraphics[width=0.22\textwidth]{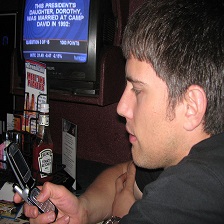} \includegraphics[width=0.22\textwidth]{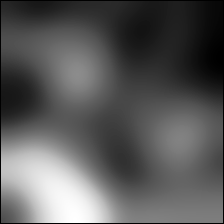} }		& 
		\thead{ \includegraphics[width=0.22\textwidth]{img_722.jpg} \includegraphics[width=0.22\textwidth]{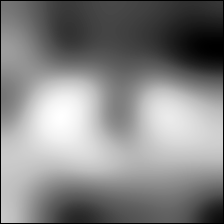} } \vspace{-0.15cm}  \\ 

		\textbf{a sandwich} sitting on top of a white plate &
		a man \textbf{sitting on a couch} on a cell phone &
		a group of people standing \textbf{around a beach} \\ 
		{\color{blue}{\textbf{a sandwich cut in half}}} on a plate &		
		a man {\color{blue}{\textbf{holding}}} a cell phone {\color{blue}{\textbf{in his hand}}} &
		a group of people standing {\color{blue}{\textbf{on a beach with surfboards}}} \\
		\textbf{a sandwich} is on a plate on a table &
		a man \textbf{sitting at a table} with a cell phone &
		a group of people standing \textbf{on top of a beach}  \vspace{0.2cm} \\ 
		
		 \thead{ \includegraphics[width=0.22\textwidth]{img_135.jpg}  \includegraphics[width=0.22\textwidth]{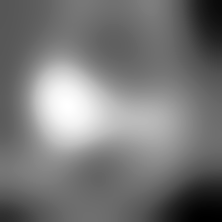}  }	& 
		\thead{ \includegraphics[width=0.22\textwidth]{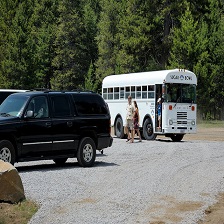} \includegraphics[width=0.22\textwidth]{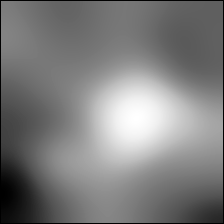} }   & 
		\thead{ \includegraphics[width=0.22\textwidth]{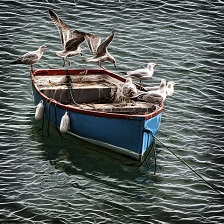} \includegraphics[width=0.22\textwidth]{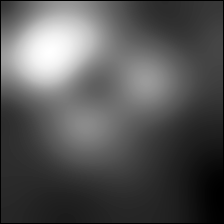} } \vspace{-0.15cm} \\ 
		
		a piece of cake sitting \textbf{on top of a plate} &	
		\textbf{a white passenger bus} is parked on a street &	
		\textbf{a boat} that is floating in the water \\
		a piece of cake {\color{blue}{\textbf{on a plate with a fork}}} &	
		{\color{blue}{\textbf{a man standing next to a bus}}} on a road  &	
		{\color{blue}{\textbf{a group of birds}}} sitting on top of a boat  \\
		a piece of chocolate cake \textbf{on a white plate}&
		\textbf{a bus} is parked on the side of the road &
		\textbf{two birds} are sitting on a dock in the water \vspace{0.2cm} \\ 

		 \thead{ \includegraphics[width=0.22\textwidth]{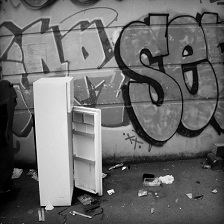}  \includegraphics[width=0.22\textwidth]{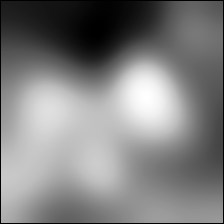}  }	& 
		\thead{ \includegraphics[width=0.22\textwidth]{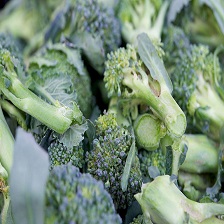} \includegraphics[width=0.22\textwidth]{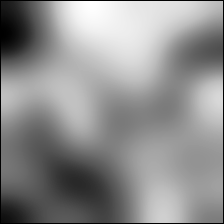} }		& 
		\thead{ \includegraphics[width=0.22\textwidth]{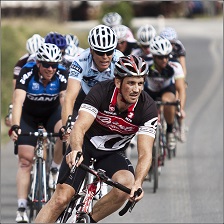} \includegraphics[width=0.22\textwidth]{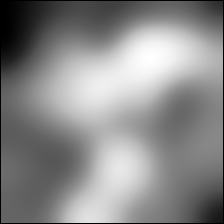} } \vspace{-0.15cm} \\ 
		
 		\textbf{a toilet} lying on its side on a sidewalk &
		\textbf{a pile of lots of broccoli} on a table &
		\textbf{a man} riding a motorcycle on the street \\
		{\color{blue}{\textbf{an old refrigerator}}} sitting in front of a wall  &		
		{\color{blue}{\textbf{a close up of a bunch of broccoli}}} &
		{\color{blue}{\textbf{a group of people}}} riding bikes down a street  \\
		a old fashioned wall with \textbf{a clock} on it &		
		\textbf{a close up of broccoli} on a table &
		\textbf{a man} riding a bike down a street  
		\end{tabular}
	}
	\caption{
		Qualitative results of our text-guided attention model.
		Two images in each example mean input image (left) and attention map (right).
		Three captions below two images represent the guidance caption (top), the generated captions from Our-ATT-kCC (middle) and Our-Uniform (bottom), respectively. 
		The more appropriate expressions generated by our attention model are marked in bold-faced blue and the corresponding phrases in the two other models are marked in bold-faced black.
	}
	\label{fig:qualitative}
\end{figure*} 

\section{Experiments}
\label{sec:experiments}
This section describes our experimental setting and presents quantitative and qualitative results of our algorithm in comparison to recent methods.

\subsection{Dataset}
\label{subsec:dataset}
We train our model on MS-COCO dataset~\cite{lin2014microsoft}, which contains 123,287 images.
The images are divided into 82,783 training images and 40,504 validation images.
Each image has five relevant captions.
We adopt the widely used splits\footnote{https://github.com/karpathy/neuraltalk} for fair comparison.
Each split of validation and testing data contains randomly selected 5,000 images from the original validation images. 
The vocabulary consists of all words that appear more than 5 times after lower-case conversion and tokenization.
The size of vocabulary amounts to 9,568 including $<$BOS$>$ and $<$EOS$>$.

\subsection{Metrics}
\label{subsec:metric}
We evaluate our model on three standard evaluation metrics, BLEU~\cite{papineni2002bleu}, METEOR~\cite{banerjee2005meteor} and CIDEr~\cite{vedantam2015cider}.
The metrics compute similarity to reference (ground-truth) captions.
BLEU computes only the precision of n-gram words between generated and reference captions.
To address the limitation of BLEU, METEOR computes both precision and recall by generating alignment between captions, where more weight is given to recall.
These two metrics are designed for automatic evaluation of machine translation methods. 
Recently, CIDEr has been introduced specifically for image captioning.
The CIDEr also measures both precision and recall by weighting n-gram words based on term frequency inverse document frequency (tf-idf).
Since this metric is used in MS COCO Captioning challenge, we mainly focus on CIDEr in our experiments.
In our experiments, we measure all performances using MS-COCO caption evaluation tool\footnote{https://github.com/tylin/coco-caption}. 

\subsection{Implementation Details}
\label{subsec:implementation}

For the image encoder, we adopt a fully convolutional network based on VGGNet (VGG-FCN), which is fine-tuned on the MS-COCO dataset to predict image attributes.
Note that this network has often been used in other image captioning methods~\cite{fang2015captions,you2016image}.
In VGG-FCN, image feature maps are obtained from the last convolutional layer given 512$\times$512 images, and our attention model is applied to the 10$\times$10 feature map.
We also test our algorithm using 101-layer Residual Network (ResNet)~\cite{He2015}, where 14$\times$14 feature maps are extracted from the layer below the global average pooling layer given 448$\times$448 images.

The entire network is trained end-to-end; we fix the parameters of two encoders, but train the text-guided attention model and the decoder from scratch.
In the decoder, the dimensionalities of the word embedding space and the hidden state of LSTM are set to 512.
We use Adam~\cite{kingma2014adam} to learn the model with mini-batch size of 80, where dropouts with 0.5 are applied to the output layer of decoder.
The learning rate starts from 0.0004 and after 10 epochs decays by the factor of 0.8 at every three epoch. 
If the decoder is trained with words in the ground-truth caption at each time step, the model often fails to generate reliable captions because the ground-truth words are unavailable in inference.  
To handle this issue, we randomly choose each input word out of the ground-truth or the prediction by the intermediate model learned at the previous iteration.
This technique is referred to as scheduled sampling~\cite{bengio2015scheduled}, which is integrated in our learning procedure after 10 epochs with ground-truth word selection probability fixed to 0.75.
Scheduled sampling makes the environment for training and inference similar and facilitates to learn more robust models.

\subsection{Performance on MS-COCO dataset}
\label{subsec:performance}

We compare our full algorithm, denoted by Ours-ATT-kCC, with three baselines and several recent algorithms.
The first baseline method (mCC) returns the caption with the most consensus.
The second approach (Ours-Uniform) is based on uniform attention, and replaces the attention layer with the average pooling layer.
The last one (Ours-ATT-mCC) is our method based only on a single guidance caption with the most consensus during inference.
As competing algorithms, we use several state-of-the-art methods, which are divided into three groups: encoder-decoder based methods (NIC~\cite{vinyals2015show}, m-RNN~\cite{mao2014deep}, LRCN~\cite{donahue2015long}, Attribute-LSTM~\cite{CVPR16what}), attention-based approaches (Semantic ATT~\cite{you2016image} and Soft- and Hard-ATT~\cite{Xu2015show}) and a composite network consisting of a series of separate subtasks (MSR~\cite{fang2015captions}).

Table~\ref{table:coco-val} summarizes the results on the test split of MS-COCO dataset.
The best performance among all algorithms in each metric is marked with bold-faced red number while the best accuracy without ResNet is marked in bold-faced blue.
Our model trained with VGG-FCN, Ours-ATT-kCC (VGG-FCN), outperforms all the compared methods in most metrics.
Interestingly, although the accuracy of retrieved captions by mCC is far from the state-of-the-art, Ours-ATT-kCC successfully learns proper attention using those captions as guidance.
Ours-ATT-kCC also outperforms Ours-Uniform and Ours-ATT-mCC consistently.
This fact shows the effectiveness of the text-guided attention and the use of multiple guidance captions in inference that makes model robust to incorrect guidance caption.
We also observe that our model improves performance substantially by adopting ResNet as the image encoder as presented in Table~\ref{table:coco-val}.

We also evaluate our models on the MS-COCO Image Captioning Challenge.
The results are presented in Table~\ref{table:coco-test}, where c5 and c40 mean the numbers of reference captions for evaluation.
In the challenge, some models improve performance by using several additional techniques such as adopting better image encoder, fine-tuning the encoder and constructing ensemble models, while our results are obtained from the identical models tested in Table~\ref{table:coco-val}.
Ours-ATT-kCC (VGG-FCN) shows outstanding results even in the condition.
Note that NIC and Semantic ATT are very competitive but rely on ensemble models while the result of ours is obtained from a single model.
Note that Ours-ATT-kCC (ResNet) still outperforms all methods with significant performance margins.

\subsection{Qualitative Analysis}
\label{subsec:qualitative}

Figure~\ref{fig:qualitative} demonstrates qualitative results for better understanding of our model, where guidance captions and attention maps are shown together.
We also compare the generated captions by our model with the uniform attention model.
The guidance caption drives our attention model to focus on relevant regions and generate detailed descriptions as shown in the first row of Figure~\ref{fig:qualitative}.
For example, our model describes a specific state of a sandwich (`sandwich cut in half') accurately while the methods relying on the guidance caption only or the caption from uniform attention model just depict it as `sandwich'.
Our model sometimes captures objects not described in guidance caption through attention as demonstrated in the second row.
The examples in the third row show that our attention model successfully attends to proper regions and generates correct caption even with noisy guidance captions.

\section{Conclusion}
\label{sec:conclusion}
We propose a text-guided attention model, where the visual attention is obtained using guidance captions.
Our model adopts exemplar-based learning approach that retrieves associated captions of visually similar images in a given training dataset.
The guidance captions highlight relevant regions and suppress unimportant ones, enabling to generate detailed captions.
To handle noise in guidance captions, we present a robust approach that exploits a set of consensus captions as guidance captions in training and testing time.
We achieve the state-of-the-art performance using our text-guided attention model on MS-COCO Captioning dataset.

\section*{Acknowledgments}
\thanks{
This work was partly supported by Samsung Electronics Co., Ltd. and the ICT R\&D program of MSIP/IITP [B0101-16-0307; ML Center, B0101-16-0552; DeepView].
}

\bibliography{bib_captioning}

\begin{thebibliography}{}

\bibitem[\protect\citeauthoryear{Andreas \bgroup et al\mbox.\egroup
  }{2016}]{andreas2015deep}
Andreas, J.; Rohrbach, M.; Darrell, T.; and Klein, D.
\newblock 2016.
\newblock Neural module networks.
\newblock In {\em CVPR}.

\bibitem[\protect\citeauthoryear{Ba, Mnih, and
  Kavukcuoglu}{2015}]{ba2014multiple}
Ba, J.; Mnih, V.; and Kavukcuoglu, K.
\newblock 2015.
\newblock Multiple object recognition with visual attention.
\newblock In {\em ICLR}.

\bibitem[\protect\citeauthoryear{Banerjee and Lavie}{2005}]{banerjee2005meteor}
Banerjee, S., and Lavie, A.
\newblock 2005.
\newblock Meteor: An automatic metric for mt evaluation with improved
  correlation with human judgments.
\newblock In {\em Proceedings of the ACL Workshop on Intrinsic and Extrinsic
  Evaluation Measures for Machine Translation and/or Summarization}.

\bibitem[\protect\citeauthoryear{Bengio \bgroup et al\mbox.\egroup
  }{2015}]{bengio2015scheduled}
Bengio, S.; Vinyals, O.; Jaitly, N.; and Shazeer, N.
\newblock 2015.
\newblock Scheduled sampling for sequence prediction with recurrent neural
  networks.
\newblock In {\em NIPS}.

\bibitem[\protect\citeauthoryear{Chung \bgroup et al\mbox.\egroup
  }{2014}]{chung2014empirical}
Chung, J.; Gulcehre, C.; Cho, K.; and Bengio, Y.
\newblock 2014.
\newblock Empirical evaluation of gated recurrent neural networks on sequence
  modeling.
\newblock In {\em Deep Learning Workshop at NIPS}.

\bibitem[\protect\citeauthoryear{Devlin \bgroup et al\mbox.\egroup
  }{2015}]{devlin2015exploring}
Devlin, J.; Gupta, S.; Girshick, R.; Mitchell, M.; and Zitnick, C.~L.
\newblock 2015.
\newblock Exploring nearest neighbor approaches for image captioning.
\newblock {\em CoRR} abs/1505.04467.

\bibitem[\protect\citeauthoryear{Donahue \bgroup et al\mbox.\egroup
  }{2015}]{donahue2015long}
Donahue, J.; Anne~Hendricks, L.; Guadarrama, S.; Rohrbach, M.; Venugopalan, S.;
  Saenko, K.; and Darrell, T.
\newblock 2015.
\newblock Long-term recurrent convolutional networks for visual recognition and
  description.
\newblock In {\em CVPR}.

\bibitem[\protect\citeauthoryear{Fang \bgroup et al\mbox.\egroup
  }{2015}]{fang2015captions}
Fang, H.; Gupta, S.; Iandola, F.; Srivastava, R.~K.; Deng, L.; Doll{\'a}r, P.;
  Gao, J.; He, X.; Mitchell, M.; Platt, J.~C.; et~al.
\newblock 2015.
\newblock From captions to visual concepts and back.
\newblock In {\em CVPR}.

\bibitem[\protect\citeauthoryear{Gregor \bgroup et al\mbox.\egroup
  }{2015}]{gregor2015draw}
Gregor, K.; Danihelka, I.; Graves, A.; and Wierstra, D.
\newblock 2015.
\newblock Draw: A recurrent neural network for image generation.
\newblock In {\em ICML}.

\bibitem[\protect\citeauthoryear{He \bgroup et al\mbox.\egroup }{2016}]{He2015}
He, K.; Zhang, X.; Ren, S.; and Sun, J.
\newblock 2016.
\newblock Deep residual learning for image recognition.
\newblock In {\em CVPR}.

\bibitem[\protect\citeauthoryear{Hong \bgroup et al\mbox.\egroup
  }{2016}]{hong2016learning}
Hong, S.; Oh, J.; Lee, H.; and Han, B.
\newblock 2016.
\newblock Learning transferrable knowledge for semantic segmentation with deep
  convolutional neural network.
\newblock In {\em CVPR}.

\bibitem[\protect\citeauthoryear{Kantorov \bgroup et al\mbox.\egroup
  }{2016}]{kantorov2016}
Kantorov, V.; Oquab, M.; Cho, M.; and Laptev, I.
\newblock 2016.
\newblock Contextlocnet: Context-aware deep network models for weakly
  supervised localization.
\newblock In {\em ECCV}.

\bibitem[\protect\citeauthoryear{Kingma and Ba}{2015}]{kingma2014adam}
Kingma, D., and Ba, J.
\newblock 2015.
\newblock Adam: A method for stochastic optimization.
\newblock In {\em ICLR}.

\bibitem[\protect\citeauthoryear{Kiros \bgroup et al\mbox.\egroup
  }{2015}]{kiros2015skip}
Kiros, R.; Zhu, Y.; Salakhutdinov, R.; Zemel, R.~S.; Torralba, A.; Urtasun, R.;
  and Fidler, S.
\newblock 2015.
\newblock Skip-thought vectors.
\newblock In {\em NIPS}.

\bibitem[\protect\citeauthoryear{Kiros, Salakhutdinov, and
  Zemel}{2015}]{kiros2014unifying}
Kiros, R.; Salakhutdinov, R.; and Zemel, R.~S.
\newblock 2015.
\newblock Unifying visual-semantic embeddings with multimodal neural language
  models.
\newblock In {\em TACL}.

\bibitem[\protect\citeauthoryear{Lin \bgroup et al\mbox.\egroup
  }{2014}]{lin2014microsoft}
Lin, T.-Y.; Maire, M.; Belongie, S.; Hays, J.; Perona, P.; Ramanan, D.;
  Doll{\'a}r, P.; and Zitnick, C.~L.
\newblock 2014.
\newblock {Microsoft COCO:} common objects in context.
\newblock In {\em ECCV}.

\bibitem[\protect\citeauthoryear{Lupyan and
  Spivey}{2010}]{lupyan2010correction}
Lupyan, G., and Spivey, M.~J.
\newblock 2010.
\newblock Making the invisible visible: Verbal but not visual cues enhance
  visual detection.
\newblock {\em PLoS ONE}.

\bibitem[\protect\citeauthoryear{Mao \bgroup et al\mbox.\egroup
  }{2015a}]{mao2015learning}
Mao, J.; Wei, X.; Yang, Y.; Wang, J.; Huang, Z.; and Yuille, A.~L.
\newblock 2015a.
\newblock Learning like a child: Fast novel visual concept learning from
  sentence descriptions of images.
\newblock In {\em ICCV}.

\bibitem[\protect\citeauthoryear{Mao \bgroup et al\mbox.\egroup
  }{2015b}]{mao2014deep}
Mao, J.; Xu, W.; Yang, Y.; Wang, J.; Huang, Z.; and Yuille, A.
\newblock 2015b.
\newblock Deep captioning with multimodal recurrent neural networks (m-rnn).
\newblock In {\em ICLR}.

\bibitem[\protect\citeauthoryear{Mishra}{2015}]{mishrainteraction}
Mishra, R.~K.
\newblock 2015.
\newblock {\em Interaction Between Attention and Language Systems in Humans}.
\newblock Springer.

\bibitem[\protect\citeauthoryear{Noh and Han}{2016}]{noh2016training}
Noh, H., and Han, B.
\newblock 2016.
\newblock Training recurrent answering units with joint loss minimization for
  vqa.
\newblock In {\em arXiv:1606.03647}.

\bibitem[\protect\citeauthoryear{Papineni \bgroup et al\mbox.\egroup
  }{2002}]{papineni2002bleu}
Papineni, K.; Roukos, S.; Ward, T.; and Zhu, W.-J.
\newblock 2002.
\newblock Bleu: A method for automatic evaluation of machine translation.
\newblock In {\em ACL}.

\bibitem[\protect\citeauthoryear{Szegedy \bgroup et al\mbox.\egroup
  }{2015}]{szegedy2015going}
Szegedy, C.; Liu, W.; Jia, Y.; Sermanet, P.; Reed, S.; Anguelov, D.; Erhan, D.;
  Vanhoucke, V.; and Rabinovich, A.
\newblock 2015.
\newblock Going deeper with convolutions.
\newblock In {\em CVPR}.

\bibitem[\protect\citeauthoryear{Vedantam, Lawrence~Zitnick, and
  Parikh}{2015}]{vedantam2015cider}
Vedantam, R.; Lawrence~Zitnick, C.; and Parikh, D.
\newblock 2015.
\newblock Cider: Consensus-based image description evaluation.
\newblock In {\em CVPR}.

\bibitem[\protect\citeauthoryear{Vinyals \bgroup et al\mbox.\egroup
  }{2015}]{vinyals2015show}
Vinyals, O.; Toshev, A.; Bengio, S.; and Erhan, D.
\newblock 2015.
\newblock Show and tell: A neural image caption generator.
\newblock In {\em CVPR}.

\bibitem[\protect\citeauthoryear{Wu \bgroup et al\mbox.\egroup
  }{2016}]{CVPR16what}
Wu, Q.; Shen, C.; Liu, L.; Dick, A.; and {van den Hengel}, A.
\newblock 2016.
\newblock What value do explicit high level concepts have in vision to language
  problems?
\newblock In {\em CVPR}.

\bibitem[\protect\citeauthoryear{Xu and Saenko}{2016}]{xu2015ask}
Xu, H., and Saenko, K.
\newblock 2016.
\newblock Ask, attend and answer: Exploring question-guided spatial attention
  for visual question answering.
\newblock In {\em VQA Challenge Workshop at CVPR}.

\bibitem[\protect\citeauthoryear{Xu \bgroup et al\mbox.\egroup
  }{2015}]{Xu2015show}
Xu, K.; Ba, J.; Kiros, R.; Cho, K.; Courville, A.; Salakhutdinov, R.; Zemel,
  R.; and Bengio, Y.
\newblock 2015.
\newblock Show, attend and tell: Neural image caption generation with visual
  attention.
\newblock In {\em ICML}.

\bibitem[\protect\citeauthoryear{Yang \bgroup et al\mbox.\egroup
  }{2016}]{yang2015stacked}
Yang, Z.; He, X.; Gao, J.; Deng, L.; and Smola, A.
\newblock 2016.
\newblock Stacked attention networks for image question answering.
\newblock In {\em CVPR}.

\bibitem[\protect\citeauthoryear{You \bgroup et al\mbox.\egroup
  }{2016}]{you2016image}
You, Q.; Jin, H.; Wang, Z.; Fang, C.; and Luo, J.
\newblock 2016.
\newblock Image captioning with semantic attention.
\newblock In {\em CVPR}.

\end{thebibliography}
\bibliographystyle{aaai} 

\end{document}